\documentclass{article}
\usepackage{ijcai25}

\usepackage{times}
\usepackage{soul}
\usepackage{url}
\usepackage[hidelinks]{hyperref}
\usepackage[utf8]{inputenc}
\usepackage[small]{caption}
\usepackage{graphicx}
\usepackage{amsmath}
\usepackage{amsthm}
\usepackage{booktabs}
\usepackage{algorithm}
\usepackage{algorithmic}
\usepackage[switch]{lineno}

\usepackage{natbib}
\usepackage{multirow} 
\usepackage{booktabs} 
\usepackage{multirow} 
\usepackage{float}  
\usepackage{subcaption}
\usepackage{adjustbox}  
\usepackage{dsfont} 
\usepackage{xcolor}

\title{LF-Steering: Latent Feature Activation Steering for Enhancing Semantic Consistency in Large Language Models}

\newcommand*{\samethanks}[1][\value{footnote}]{\footnotemark[#1]}

\author{
Jingyuan Yang$^{1,2}$\and
Rongjun Li$^2$\and
Weixuan Wang$^3$\and
Ziyu Zhou$^2$\\
Zhiyong Feng$^1$\thanks{Corresponding author.}\And
Wei Peng$^2$\samethanks \\
\affiliations
$^1$College of Intelligence and Computing, Tianjin University\\
$^2$IT Innovation and Research Center, Huawei Technologies\\
$^3$Informatics, University of Edinburgh\\
\emails
\{yangjingyuan2, lirongjun3, zhouziyu8, peng.wei1@huawei.com\}@huawei.com,
weixuan.wang@ed.ac.uk,
zyfeng@tju.edu.cn
}

\begin{document}

\maketitle

\begin{abstract}
Large Language Models (LLMs) often generate inconsistent responses when prompted with semantically equivalent paraphrased inputs. Recently, activation steering, a technique that modulates LLMs' behaviours by adjusting their latent representations during inference time, has been explored to improve the semantic consistency of LLMs. However, these methods typically operate at the model component level, such as layer hidden states or attention head outputs. They face a challenge due to the ``polysemanticity issue'', where the model components of LLMs typically encode multiple entangled features, making precise steering difficult. To address this challenge, we drill down to feature-level representations and propose \textit{LF-Steering}, a novel activation steering approach to precisely identify latent feature representations responsible for semantic inconsistency. More specifically, our method maps the hidden states of the relevant transformer layer into a sparsely activated, high-dimensional feature space based on a sparse autoencoder (SAE), ensuring model steering based on decoupled feature representations with minimal interference. Comprehensive experiments on NLU and NLG datasets demonstrate the effectiveness of our method in enhancing semantic consistency, resulting in significant performance gains for various NLU and NLG tasks.
\end{abstract}

\begin{figure}[ht]
    \centering  
    \includegraphics[width=0.67\linewidth]{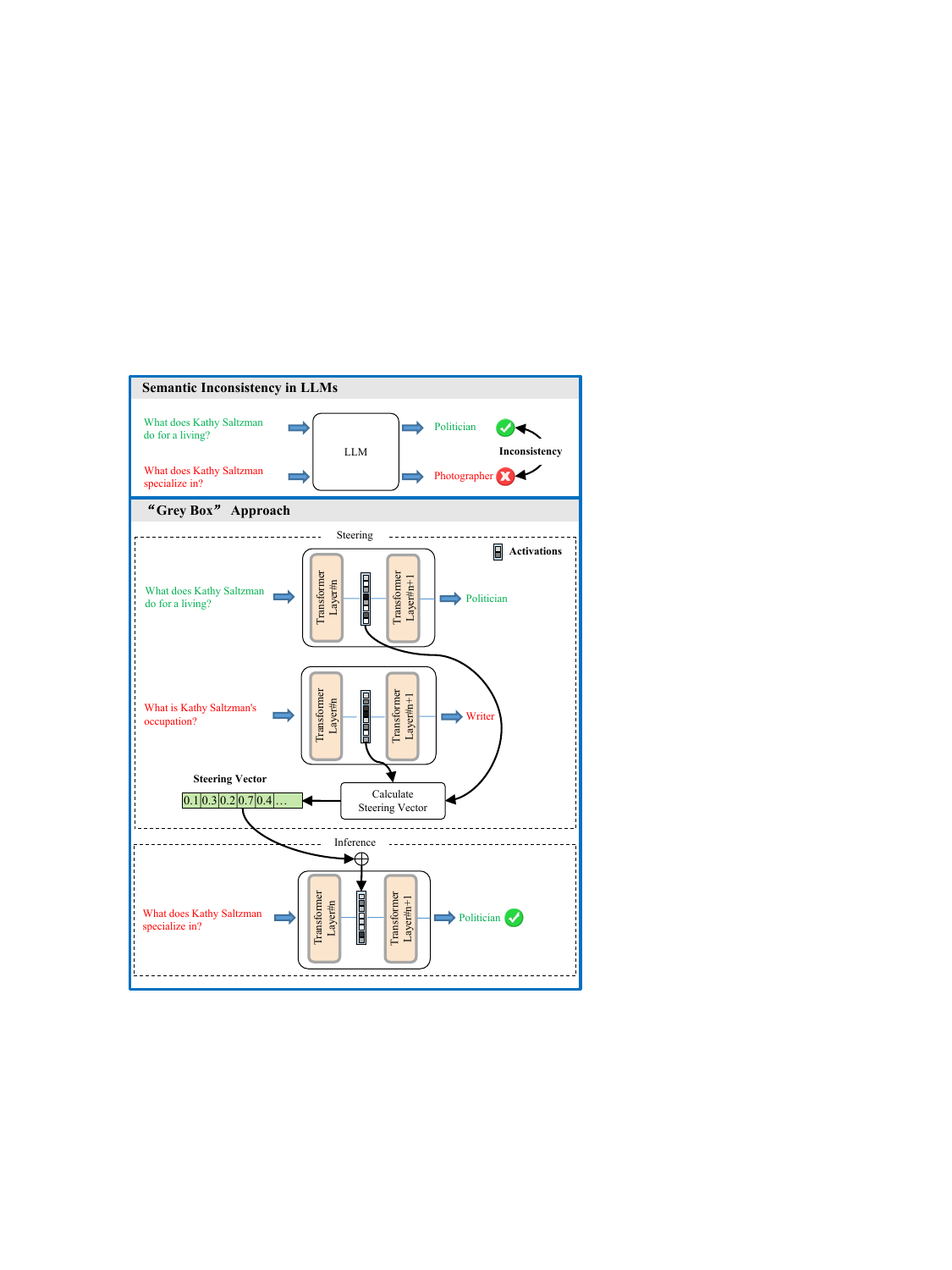}
    \caption{Semantic inconsistency in LLMs and how a ``grey-box'' approach based on activation steering addresses this problem.}
    \label{fig:intro}
\end{figure}
\section{Introduction}

Despite the remarkable capabilities demonstrated by Large Language Models (LLMs) across a wide range of tasks, they often exhibit a tendency to produce inconsistent responses when presented with semantically equivalent inputs phrased differently. This phenomenon, commonly referred to as "semantic inconsistency" \citep{gan2023sensitivity, rabinovich2023predicting, raj2022measuring,wang-etal-2024-assessing,  fierro2024does}, poses a significant challenge to the practical deployment of LLMs in real-world applications.

Existing approaches to improve semantic consistency for LLMs can be broadly classified into two categories: end-to-end ``black-box'' methods and interpretability-oriented ``grey-box'' approaches with greater transparency (Figure \ref{fig:intro}). In ``black-box'' methods, data-driven supervised fine-tuning (SFT) \citep{ouyang2022training, fierro2024does, zhao-etal-2024-improving} is commonly employed, using pairs of prompts and model responses with semantically equivalent meanings as training data. While these methods have shown effectiveness, they come with significant costs in data engineering and LLM fine-tuning.  Recently, \citet{yang-etal-2024-enhancing} propose a ``grey-box'' method aimed at improving the semantic consistency of LLMs from a more transparent perspective. This approach identifies specific internal representations (e.g., attention head outputs) responsible for semantic inconsistency, and subsequently adjusts them toward a more semantically consistent direction to steer model behaviour. However, due to the ``polysemanticity issue'' \citep{elhage2022toy}, individual model components (i.e., neurons, attention heads, etc.) of an LLM typically encode a mixture of multiple unrelated features, making it challenging to precisely adjust model behaviours based on these entangled representations.

To address this challenge, we drill down to finer-grained feature-level representations of LLMs and propose a novel approach named ``\textit{LF-Steering}'' to precisely identify and modify the relevant latent feature representations responsible for semantic inconsistency. The features are constructed by decomposing transformer layer representations into a sparsely activated, higher-dimensional feature space. This ensures that the model steering is based on decoupled feature representations with minimal interference. Our approach involves the following key steps. First, we apply a sparse autoencoder (SAE) to map the hidden states of the transformer layers into a feature space, and then identify relevant features that significantly impact on the LLM's semantic consistency. After that, we selectively modify the identified key features to improve the model's semantic consistency.

As demonstrated by the experiments, our proposed method achieves state-of-the-art (SOTA) semantic consistency scores, leading to substantial performance gains over baselines across a wide range of NLU and NLG tasks. In summary, our contributions are two-fold:

\begin{itemize}
    \item We propose a novel latent feature activation steering method that decomposes a model's internal representations into a sparsely activated, higher-dimensional feature space. This enables model steering at finer-grained feature representations to enhance LLMs' semantic consistency. 
    \item Our proposed method is a generic activation steering technique to address semantic inconsistency across a wide range of tasks. Extensive experiments demonstrate its effectiveness, achieving SOTA performance over existing leading activation steering methods in enhancing semantic consistency of LLMs. It outperforms the baselines in prediction accuracy by up to 8.9\% on NLU datasets. For NLG datasets, accuracy can be improved by up to 7.12\%. 
\end{itemize}

\section{Related Work}
\noindent \textbf{Semantic Consistency.}
LLMs tend to produce semantic inconsistent results \citep{gan2023sensitivity, fierro2024does, wang-etal-2024-assessing} when prompted with semantically equivalent paraphrased inputs. Current methods for improving semantic consistency of LLMs fall into two categories: end-to-end ``black-box'' methods and interpretability-oriented ``grey-box'' approaches. ``Black-box'' methods primarily relied on expert-designed prompting strategies \citep{raj2023semantic} or SFT techniques \citep{ouyang2022training, zhou-etal-2022-prompt, fierro2024does, zhao-etal-2024-improving}. Among them, \citet{raj2023semantic} leveraged an Ask-to-Choose (A2C) prompting technique to improve the accuracy as well as the semantic consistency of LLMs. Additionally, \citet{zhao-etal-2024-improving} proposed a data augmentation method to generate synthetic data consisting of pairs of paraphrased prompts and LLM responses for SFT, aiming to achieve the same end. Recently, \citet{yang-etal-2024-enhancing}  attempted to address this issue through a more transparent ``grey-box'' method. \citet{yang-etal-2024-enhancing} applied an activation steering technique to identify specific internal hidden state representations (such as attention head outputs) contributing to semantic inconsistency and subsequently adjusted these representations to achieve greater semantic consistency. Despite its effectiveness, the method was limited to operating on model components subjective to the aforementioned ``polysemanticity issue''.

In contrast, our steering method drills down to finer-grained feature units, enabling more precise steering and achieving superior performance compared to existing methods.
\\ \\
\noindent  \textbf{Activation Steering.} Activation steering techniques are emerged to modify specific model behaviors by adjusting their internal representations (i.e., the activation values of neurons). Existing activation steering approaches mostly rely on the hidden states of transformer layers. For instance,
\citet{DBLP:journals/corr/abs-2308-10248} introduced an activation addition approach, which computes a steering vector from contrastive samples to guide model behaviours, such as generating specific sentiments or topics in its responses. \citet{DBLP:conf/acl/RimskyGSTHT24} employed mass mean activation difference to improve model steering performance. Other methods operated on the outputs of attention heads to steer model behaviors. For example, \citet{li2023inference} utilised linear probing to identify the attention heads correlated with truthful responses. By adjusting the outputs of these attention heads, they enhanced the truthfulness of an LLM's responses.  \citet{DBLP:conf/aaai/ChenSJLKWX24} proposed a method to enhance the truthfulness of an LLM's responses by modifying hidden state outputs along multiple orthogonal truthful directions identified by a linear probe.

Unlike previous steering methods that relied on either layer-wise or attention-head outputs, our approach operates at the feature level with the aim to achieve more precise activation steering.
\\ \\
\noindent \textbf{Sparse Autoencoder.}
The study of sparse autoencoders (SAEs) was first introduced by \citet{agarwal2016learning} as an unsupervised learning method for feature extraction. More recently, SAEs have been used to disentangle representations within LLMs for analysing polysemanticity phenomena in these models \citep{bricken2023towards}. For example, \citet{cunningham2023sparse} trained SAEs with L1 loss to map the internal activations of LLMs into a higher-dimensional feature space, using these feature representations to interpret model behaviors. More recently, gated sparse autoencoder \citep{rajamanoharan2024improving} was proposed to address the imbalance between accurate reconstruction and activation sparsity caused by L1 loss biases. \citet{gao2024scaling} further proposed TopK SAE, leveraging a Top-K activation function to enforce sparsity constraints more effectively to this end.

In this paper, we adopt a TopK SAE to decompose transformer layer representations into a sparsely activated, higher-dimensional feature space, ensuring that the features are decoupled with minimal interference for precise activation steering.

\begin{figure*}[ht]
    \centering
    \includegraphics[width=0.93\textwidth]{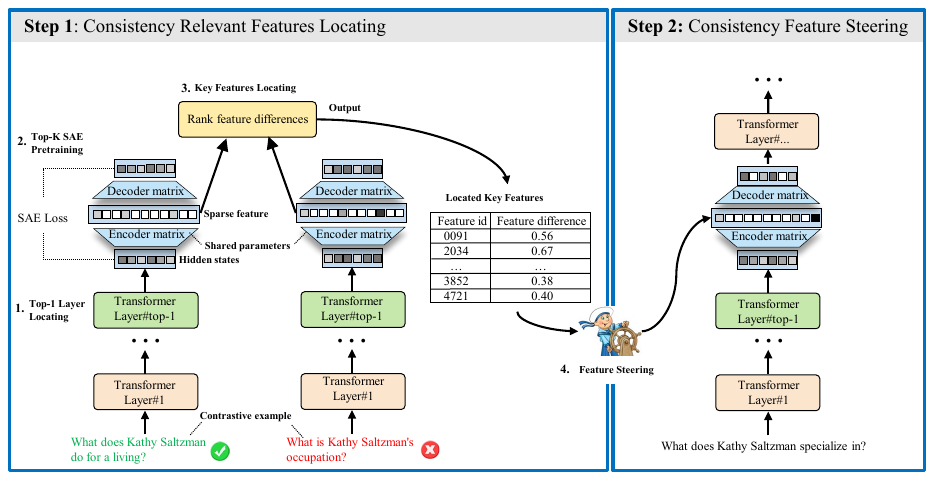}
    \caption{The flowchart of our method. (1) Consistency Relevant Features Locating consists two steps. We first use the method proposed by \citep{yang-etal-2024-enhancing} to locate the top-1 transformer layer and then pretrain a SAE and use it to locate the key features responsible for the semantic inconsistencies in the LLM. (2) Steering the LLM towards greater semantic consistency by adjusting the values of the identified key features.}
    \label{fig:my_image}
\end{figure*}

\section{Methodology}
As illustrated in Figure \ref{fig:my_image}, our proposed method consists of two main steps: \textit{Consistency Relevant Features Locating} and \textit{Consistency Feature Steering}. In the first step, we employ a coarse-to-fine locating strategy. Initially, we use the method proposed by \citet{yang-etal-2024-enhancing} to identify the top-1 transformer layer that has the most significant impact on the model's semantic inconsistency. Subsequently, we pretrain a spare autoencoder (SAE) and use it to project the hidden states of the identified transformer layer into a higher-dimensional feature space. Within this feature space, we locate the key features that are most relevant to the LLM's semantic inconsistency. In the second step, we enable the LLM to produce more semantically consistent responses by adjusting the values of the identified key features.

\subsection{Consistency Relevant Features Locating}
\label{sec:layer_locating}
\textbf{Top-1 Transformer Layer Locating.} We use the method proposed by \citet{yang-etal-2024-enhancing} to locate the top-1 transformer layer influences the LLM's semantic consistency. Specifically, the layer locating dataset $\mathcal{D}_{l}$ was first built, comprising consistent pairs in the form $([m,n],l)$, where $m$ is the input prompt and $n$ is a rephrased version of $m$, which can be generated using large-scale LLMs like GPT-4\footnote{https://platform.openai.com/docs/api-reference/chat}. The variable $l$ denotes whether the LLM outputs are consistent for $m$ and $n$, which can also be evaluated by large-scale LLMs. Following \citet{yang-etal-2024-enhancing}, the $\mathcal{D}_{l}$ dataset, consisting of 500 instances, was randomly split into a training set and a test set with a 4:1 ratio. Next, for each transformer layer, we train a classifier that takes the concatenated hidden states of the corresponding layer for prompts $m$ and $n$ as inputs and uses the consistency label $l$ as the ground truth. 

High classification accuracy on the test set indicates that the layer has a greater influence on the LLM's semantic inconsistency, while low accuracy suggests that the layer is less relevant to semantic inconsistency. The top-1 performing transformer layer is selected by ranking the test set classification accuracy.
\\ \\
\noindent \textbf{TopK SAE Pretraining.} Since the hidden states of the transformer layers contain entangled feature representations \citep{elhage2022toy}, this hinders the precise locating of features related to semantic consistency. To address this, it is essential to delve into finer-grained feature-level representations for accurate feature identification. 

To achieve this, we pretrain a TopK SAE \citep{gao2024scaling} and use it to decompose the dentified top-1 transformer layer representations into a higher-dimensional, sparsely activated feature space, resulting in decoupled feature-level representations. 

More specifically, the TopK SAE consists of two main components: the SAE encoder and the SAE decoder. The encoder maps the model's hidden states into a high-dimensional feature space and applies the TopK activation function to retain the $k$ most significant latent features, ensuring sparse activation. The decoder then reconstructs the original hidden states from these top $k$ latent features. The pretraining process is formally represented as follows:

\begin{equation}
\begin{gathered}
    z(d) = \text{TopK}(W_{\text{enc}} (h_\text{top-1}(d) - b_{\text{pre}}))\\
    \hat{h}_\text{top-1}(d) = W_{\text{dec}} z(d) + b_{\text{pre}}, \\
    \mathcal{L} = \frac{1}{| \mathcal{D}_{o} |} \sum_{d}\| h_\text{top-1}(d) - \hat{h}_\text{top-1}(d)\|_2^2, d \in \mathcal{D}_{o}. \\
\end{gathered}
\end{equation}

Here, $d$ is the data instance from the open domain dataset $\mathcal{D}_{o}$. \( h_\text{top-1}(d) \) and \( \hat{h}_\text{top-1}(d) \) denote the hidden states and the reconstructed hidden states from the identified top-1 transformer layer respectively. \( W_{\text{enc}} \), \( W_{\text{dec}} \), and \( b_{\text{pre}} \) represent the encoder weights, decoder weights, and the corresponding bias term, respectively.  The \( \mathcal{L}\) is the training loss. 

Due to the GPU memory limitations, the TopK SAE \citep{gao2024scaling} used in our paper decomposes the 4,096 dimensional hidden states to 65,536 dimensional features. We select the top 192 latent features to recover the original hidden states. We leverage 850K public data $\mathcal{D}_{o}$, collected from various domains \footnote{https://huggingface.co/datasets/togethercomputer/RedPajama-Data-1T-Sample} to pretrain the TopK SAE. The reason we do not use task-specific data is to ensure that the learned feature representations remain generalizable. The training process is performed with a batch size of 1 per GPU, utilizing 8xV100 GPUs each with 32 GB of memory. We train the TopK SAE for a total of 32,000 steps with the learning rate is set to 2e-4.
\\ \\
\noindent \textbf{Key Features Locating.} To identify key features related to semantic inconsistency within the feature space decomposed from the pretrained SAE, we build dataset $\mathcal{D}_\text{f}$ consists of 500 contrastive examples in the form $([u,v])$, where $u$ represents the input prompt and $v$ is its rephrased version obtained from GPT-4. The key difference is that $u$ generates a correct result when processed by the LLM, whereas $v$ produces an erroneous response from the target LLM. The correctness labels are assigned based on whether their predicted outputs are identical for NLU tasks. For NLG tasks, GPT-4 is also used to evaluate the correctness label.

Next, we feed the contrastive examples $u$ and $v$ from $D_\text{f}$ into the LLM to obtain their respective last token feature representations from the pretrained SAE. We then select key features related to the LLM's semantic inconsistency by selecting average feature differences from contrastive examples that exceed a specified threshold value $t$. This process is formally represented as follows:
\begin{equation}
\begin{gathered}
   z(u) = \text{TopK}(W_{\text{enc}}( h_\text{top-1}(u) - b_{\text{pre}})), \\
   z(v) = \text{TopK}(W_{\text{enc}}( h_\text{top-1}(v) - b_{\text{pre}})), \\
   g = \frac{1}{| \mathcal{D}_\text{f} |} \sum_{(u,v)}|z(u)-z(v)|, (u, v) \in \mathcal{D}_\text{f}, \\
   \mathcal{I} = \{ i \mid |g_i - t| > 0, i =1,2,\dots, |g| \}. 
   \label{eq:loc_feats}
\end{gathered}
\end{equation}
Here, $z(u)$ and $z(v)$ are the last token feature activations of $u$ and $v$ respectively. $g$ is the average feature differences, and $|g|$ is the size of $g$. $\mathcal{I}$ contains the indexes of the located key features. $t$ is a predefined feature difference threshold.

\subsection{Consistency Feature Steering}
In this step, we make precise feature representations adjustments to enhance semantic consistency. Specifically, when an input prompt ($q$) is provided during inference, we check whether its activated features indexes from the pretrained TopK SAE are present in the identified key feature indexes $\mathcal{I}$ (See equation \ref{eq:loc_feats}). If so, we add the corresponding consistency feature bias (average feature differences) on it to guide the LLM towards greater semantic consistency. Formally, this steering process is calculated as follows:

\begin{equation}
\begin{gathered}
  z(q) = \text{TopK}(W_{\text{enc}} (h_\text{top-1}(q) - b_{\text{pre}})), \\
  b_{i} = \mathds{1}(i \in \mathcal{I} \  \text{and} \ z_{i}(q) \neq 0 ) \cdot g_i \\
  z_{i}(q) = z_{i}(q) + \alpha \cdot  b_{i}, i =1,2,\dots, |z|.
\end{gathered}
\end{equation}
Here, $q$ is the given input prompt, $z(q)$ is the activated features from the pretrained TopK SAE. $\mathds{1}$ denotes the indicator function, which represents the intersection of the located key feature indexes $\mathcal{I}$ and the activated feature indexes of $z(q)$. $b_i$ is the consistency feature bias (average feature differences). $\alpha$ is a hyperparameter that controls the strength of steering. 

\section{Experiments}
\subsection{Datasets and Evaluation Metrics}
To ensure a fair comparison, we follow the same evaluation framework used by \citet{yang-etal-2024-enhancing}. This includes using the RobustBOOLQ, RobustSST2, and RobustMRPC datasets for NLU tasks, and the PopQA\_Sport and PopQA\_capital datasets for NLG tasks. For task performance evaluation, the overall output accuracy across diverse paraphrased inputs are employed. To assess the semantic consistency of the LLM, we use the standard deviation of the overall output accuracy and the mean pairwise cosine similarity for NLU and NLG datasets respectively.

\begin{table*}[htbp]
\centering
\begin{tabular}{@{}l@{  }c@{  }c@{  }c@{  }c@{  }c@{}}
\toprule
\textbf{Method} & \textbf{RobustBOOLQ} & \textbf{RobustSST2} & \textbf{RobustMRPC} & \textbf{PopQA\_Sport} & \textbf{PopQA\_Capital} \\ \midrule
LLama2-7B-Chat & $46.40_{ \pm 10.55}$ & $85.66_{ \pm 4.88}$ & $67.15_{ \pm 5.36}$ & $50.83_{/0.79}$ & $73.33_{/0.73}$ \\
+ActAdd & $55.70_{ \pm 11.63}$ & $\mathbf{89.44_{ \pm 4.51}}$ & $49.50_{ \pm 17.44}$ & $\mathbf{57.31_{/0.79}}$ & $73.83_{/0.79}$ \\ 
+CAA & $53.90_{ \pm 9.58}$ & $86.46_{ \pm 4.71}$ & $68.38_{ \pm 4.77}$ & $53.78_{/0.79}$ & $\mathbf{75.03_{/0.81}}$ \\ 
+SCS & $\mathbf{57.50_{ \pm 5.10}}$ & $89.90_{ \pm 4.54}$ & $\mathbf{68.62_{ \pm 4.47}}$ & $53.20_{/0.80}$ & $74.36_{/0.77}$ \\ 
\midrule
+LF-Steering & $\mathbf{66.40_{ \pm 3.39}}$ & $\mathbf{90.13_{ \pm 3.14}}$ & $\mathbf{68.38_{ \pm 4.41}}$ & $\mathbf{64.43_{/0.78}}$ & $\mathbf{75.19_{/0.83}}$ \\ 
\bottomrule
\end{tabular}
\caption{The main experiment results on the NLU and NLG datasets. The $46.40_{\pm10.55}$ notation shows an average test set accuracy of $46.40$, with a standard deviation of $10.55$, whereas $73.33_{/0.73}$ indicates an average test set accuracy of $73.33$ and a mean pairwise cosine similarity of $0.73$.} 

\label{tab:acc_combined}
\end{table*}

\subsection{Baseline Methods}
In this paper, we compare three baseline methods on the top of our LLM backbone LLama2-7B-Chat. The three model component-level activation steering methods are based on the hidden states of transformer layer and attention head outputs.\\

\noindent \textbf{Backbone LLama2-7B-Chat} \citep{touvron2023llama} is an advanced auto-regressive language model\footnote{https://huggingface.co/meta-llama/Llama-2-7b-chat-hf} with 7 billion parameters, pretrained and post-trained on massive datasets.\\

\noindent \textbf{ActAdd} \citep{DBLP:journals/corr/abs-2308-10248} computed a steering vector based on the transformer layer hidden states from one contrastive sample to guide model behavior. Building on this approach, we randomly select one contrastive example from the corresponding dataset and utilize the hidden states from the same top-1 identified transformer layer in our paper to compute the steering vector in our experiments.\\

\noindent \textbf{CAA} \citep{DBLP:conf/acl/RimskyGSTHT24} utilized mean difference layer representations from contrastive samples to steer model behavior\footnote{https://github.com/nrimsky/CAA}. Following this method, we use the same amount of training data to compute the mass mean steering vector from the top-1 identified transformer layer and the same corresponding dataset.\\

\noindent \textbf{SCS}
\citep{yang-etal-2024-enhancing} introduced an activation steering technique that identifies key components, such as attention heads, that influence semantic consistency. Then, it adjusted the representations of these attention heads to guide the LLM towards greater semantic consistency, thereby improving its overall performance. In this paper, we use the experimental results from their study. Note that, since no specific method name was given to this method in the original work, we use Semantic Consistency Steering (SCS) to refer to this method for clarity in subsequent experiments.

\subsection{Main Experimental Results}
As shown in Table \ref{tab:acc_combined}, our proposed activation steering method demonstrates significant improvements in both semantic consistency and task performance across a range of NLU and NLG datasets. 

For NLU datasets, our approach outperforms the original LLama2-7B-Chat model by achieving an average reduction of 3.28\% in standard deviation and an 8.56\% improvement in prediction accuracy. When compared to the previous SOTA activation steering method, our method reduces standard deviation by up to 1.71\% and increases prediction accuracy by up to 8.9\%. 

For NLG datasets, our approach improves semantic consistency scores by an average of 4.5\% and accuracy by 7.73\% compared to the original LLama2-7B-Chat. Against the previous SOTA activation steering methods, our method yields up to 2.0\% increase in average semantic consistency scores and up to 7.12\% rise in accuracy.

\subsection{Top-1 Transformer Layer Locating Results}
\begin{figure}[htbp]
    \centering
    \includegraphics[width=0.95\columnwidth]{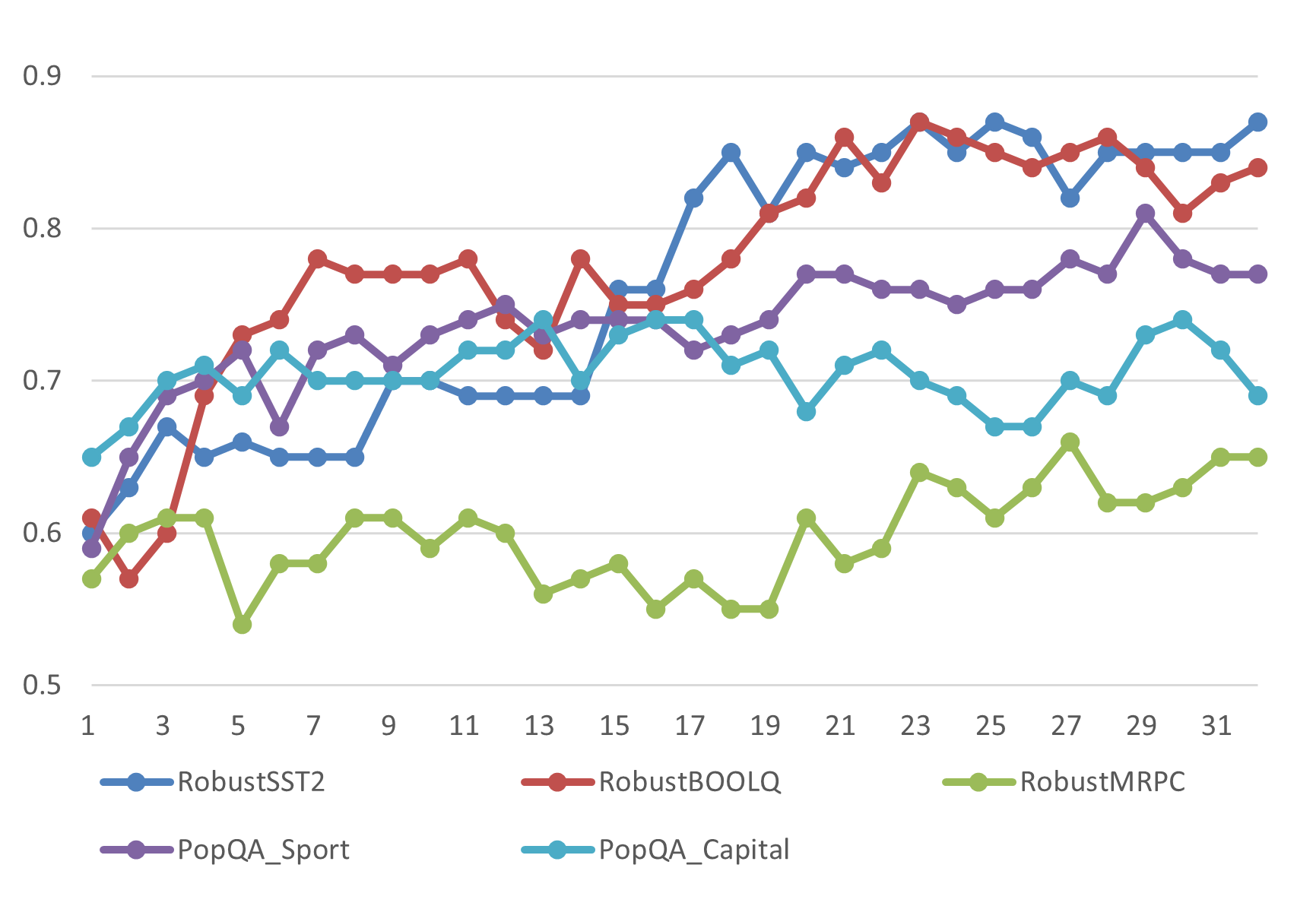}
    \caption{Comparison of layer-wise locating accuracy across 32 transformer layers for the experiment datasets.}
    \label{fig:layer_locating2}
\end{figure}

\begin{table*}[htbp]\footnotesize
\centering
    \begin{adjustbox}{width=\textwidth}
    \begin{tabular}{l l l l}
    \toprule
    \textbf{Datasets} & \textbf{Feature IDs} & \textbf{Interpretation} & \textbf{Examples}  \\ 
    \midrule
    \multirow{3}{*}{RobustBOOLQ} & \multirow{3}{*}{15780} & \multirow{3}{*}{Looking for Answer} & ...hold the key to the question\textcolor{orange}{?} Kindly \textcolor{orange}{answer} 'yes' or 'no'...      \\
                           &          &           & ...provides the answer to the question\textcolor{orange}{? Say} 'yes' or 'no'...     \\
                           &          &           & ...passage answers the question\textcolor{orange}{? Respond} with 'yes' or 'no'... \\
   \midrule
    \multirow{3}{*}{PopQA\_Capital} & \multirow{3}{*}{26247} & \multirow{3}{*}{Location-related questions} & \textcolor{orange}{What} is \textcolor{orange}{Ireland}'s capital\textcolor{orange}{?}      \\
                           &           &          & \textcolor{orange}{The} capital city of \textcolor{orange}{Ireland} is what\textcolor{orange}{?}     \\
                           &        &         & \textcolor{orange}{What} is the capital of \textcolor{orange}{North Carolina?}      \\
    \bottomrule
    \end{tabular}
    \end{adjustbox}
\caption{Located key SAE features with manually summarized interpretations and corresponding example activations (highlighted in orange) for the NLU (RobustBOOLQ) and NLG (PopQA\_Capital) datasets.} 
\label{tab:feat_inter}
\end{table*}

\begin{table*}[ht]
\centering
\begin{tabular}{lccccc}
\hline
\textbf{Method} & \textbf{RobustBOOLQ} & \textbf{RobustSST2} & \textbf{RobustMRPC} & \textbf{PopQA\_Sport} & \textbf{PopQA\_Capital} \\
\hline
LLaMA2-7B-Chat & $46.40_{ \pm 10.55}$ & $85.66_{ \pm 4.88}$ & $67.15_{ \pm 5.36}$ & $50.83_{ / 0.79}$ & $73.33_{ / 0.73}$ \\
+LF-Steering (neuron-level) & $50.50_{ \pm 9.07}$ & $86.35_{ \pm 4.78}$ & $68.13_{ \pm 4.91}$ & $52.74_{ / 0.79}$ & $74.36_{ / 0.76}$ \\
+LF-Steering (feature-level) & $\mathbf{66.40_{ \pm 3.39}}$ & $\mathbf{90.13_{ \pm 3.14}}$ & $\mathbf{68.38_{ \pm 4.41}}$ & $\mathbf{64.43_{ / 0.78}}$ & $\mathbf{75.19_{ / 0.83}}$ \\
\hline
\end{tabular}
\caption{Comparison of feature-level and neuron-level LF-Steering methods for enhancing LLM's semantic consistency and task performance.}
\label{tab:neuron_feature_compare}
\end{table*}

We use the LLama2-7B-Chat model and the top-1 transformer layer locating method mentioned in Section \ref{sec:layer_locating} to analyze the contributions of transformer layers to the model’s semantic consistency. As shown in Figure \ref{fig:layer_locating2}, we observe a notable trend: the locating accuracy is relatively higher between layers 17 and 32, indicating that the middle to final layers are significantly associated with semantic consistency. Notably, the high locating accuracy in the final layers is evident. It appears these layers may relate to the model's decision-making process and thus playing a more critical role for steering the final response. For the mid-layers, discrepancies in semantic representations for paraphrased inputs may lead to variations in subsequent prediction outcomes. In contrast, the earlier layers primarily capture syntactic and literal information \citep{DBLP:conf/acl/LiuKL024}, making it challenging to accurately assess the consistency of their semantic representations; consequently, the locating accuracy is relatively low, hovering around 65\%.

\subsection{Interpretation of Feature Representations}
As shown in Table \ref{tab:feat_inter}, we select several examples to demonstrate the pattern captured by the located key SAE features and explain how they may contribute to enhancing the model's semantic consistency. Specifically, we choose the RobustBOOLQ dataset for NLU and the PopQA\_Capital dataset for NLG. We randomly select the top-ranked feature IDs and manually inspect and interpret them using the corresponding datasets. It is found that feature ID 15780 in the RobustBOOLQ dataset is related to ``looking for answers'', while feature ID 26247 in the PopQA\_Capital dataset is linked to ``location-related questions''. Enhancing the activations of these features improves the model’s alignment with various semantically equivalent expressions, while exerting lower interference on other features, thereby resulting in improved performance in enhancing semantic consistency.

\subsection{Comparing the Effectiveness of Feature-Level and Neuron-Level Steering}
To validate the effectiveness of our proposed feature-level steering method compared to neuron-level steering for enhancing LLM's semantic consistency, we designed a comparative experiment. In specific, LF-Steering (neuron-level) steer the identified top-1 transformer hidden states to enhance LLM's semantic consistency, whereas LF-Steering (feature-level) steer the SAE features.

As shown in Table \ref{tab:neuron_feature_compare}, LF-Steering (features) significantly outperforms LF-Steering (hidden states), achieving an average accuracy increase of 6.64\% and an average standard deviation reduction of 2.6 across NLU datasets. Additionally, it achieves an average accuracy increase of 6.26\% and enhances the average pairwise cosine similarity score by 3\% across NLG datasets. These results highlight the superior advantages of our proposed feature-level steering approach in enhancing the task performance and semantic consistency of LLMs.

\subsection{The Effectiveness of Transformer Layer and SAE Feature Locating}
To investigate the effectiveness of the transformer layer and the SAE feature locating in our steering process, we conduct two experiments. The first experiment involves randomly selecting a transformer layer for steering, while the second focuses on randomly choosing SAE features for modification.  

\begin{table}[htbp]
\centering
\begin{tabular}{@{}lcc@{}}
\toprule
\textbf{Method}  & \textbf{RobustBOOLQ}  \\ \midrule
LLama2-7B-Chat & $46.50_{ \pm 10.55}$ \\ 
+LF-Steering & $66.40_{ \pm 3.39}$ \\ 
w/ random layer & $58.40_{ \pm 8.01}$ \\
w/ random features & $51.10_{ \pm 16.59}$ \\ 
\bottomrule
\end{tabular}
\caption{Ablation studies on the influence of model layer and SAE features locating strategy. The term ``random layer'' refers to  randomly selecting a transformer layer, while ``random features'' refers to randomly selecting SAE features.} 
\label{tab:abl_head}
\end{table}

The results shown in Table \ref{tab:abl_head} demonstrate that randomly selecting the transformer layer or features for steering leads to decreased semantic consistency and task performance on the RobustBOOLQ dataset. In comparison to our proposed steering method, the random selection of transformer layer results in a decrease of 4.62\% in semantic consistency (as measured by the standard deviation of accuracy) and an 8.0\% decline in accuracy. Similarly, randomly selecting SAE features leads to a more pronounced decrease, with semantic consistency decreased by 13.20\% and accuracy reduced by 15.30\%. These findings underscore the critical importance of selecting an appropriate transformer layer and SAE features for effective activation steering, demonstrating the superiority of our proposed method over random selection approaches.

\subsection{Analysis of Hyperparameters}
\subsubsection{The Impact of Feature Difference Threshold}  
\begin{figure}[htbp]
    \centering
     \includegraphics[width=\columnwidth]{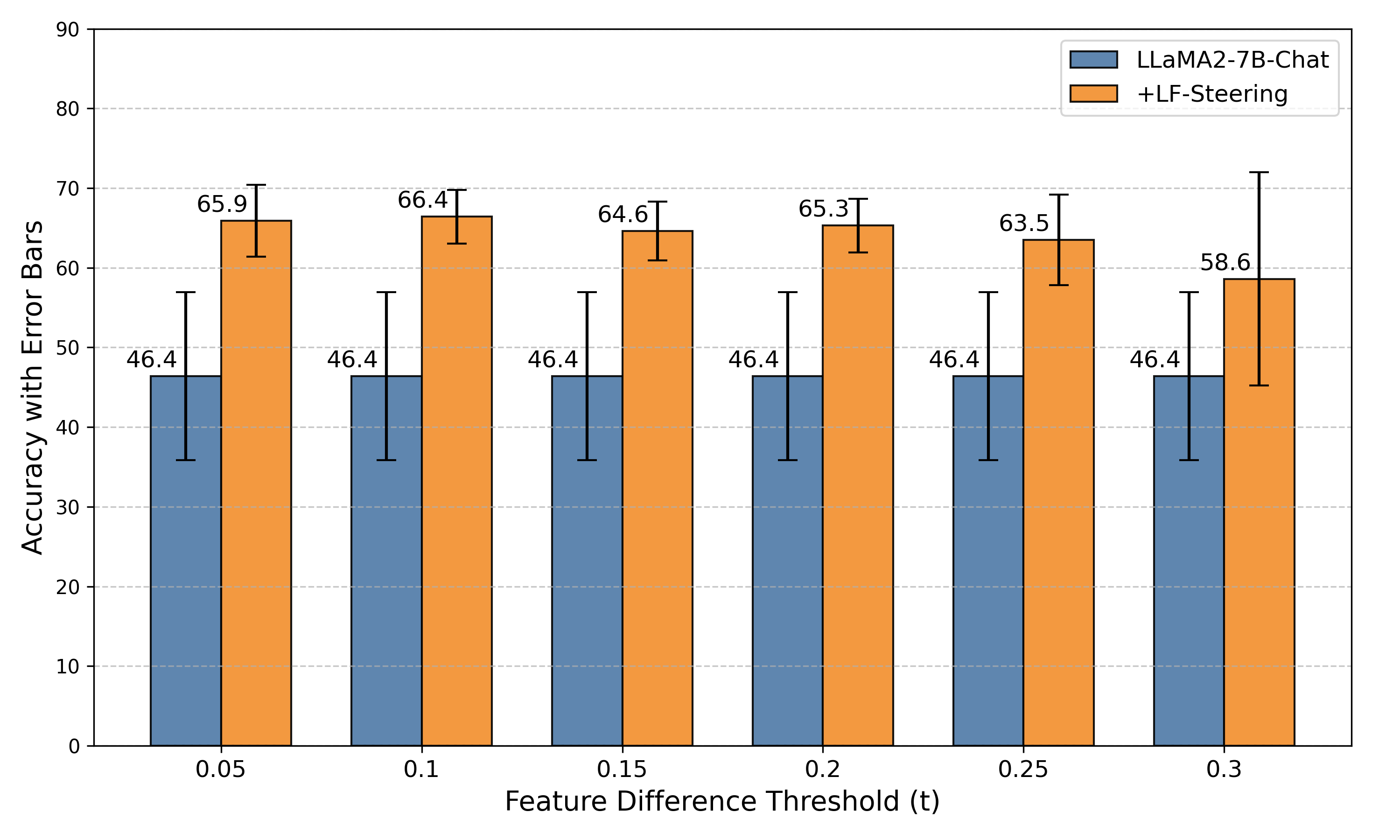}
    \caption{Performance of our proposed activation steering method across different threshold values.}
    \label{fig:abl_params}
\end{figure}

We conduct an analysis of the impact of varying feature difference thresholds in our experimental setup using the RobustBOOLQ dataset. We examine the feature difference threshold $t$ in the set of $\{0.05, 0.10, 0.15, 0.20, 0.25, 0.30\}$. As shown in Figure \ref{fig:abl_params}, the steered model achieves peak accuracy when the $t$ is approximately 0.10. However, we observe a notable decline in predicting accuracy coupled with an increase in standard deviation when the threshold $t$ reaches or exceeds 0.25. This performance decrease can be attributed to an excessively high feature difference threshold, which limits the number of steered features thus results in insufficient steering. These findings emphasize the critical nature of selecting an optimal threshold $t$ for effective steering. In this paper, we set $t$ to a value of $0.1$.

\subsubsection{The Impact of Feature Activation Strength}
\begin{figure}[htbp]
    \centering
     \includegraphics[width=\columnwidth]{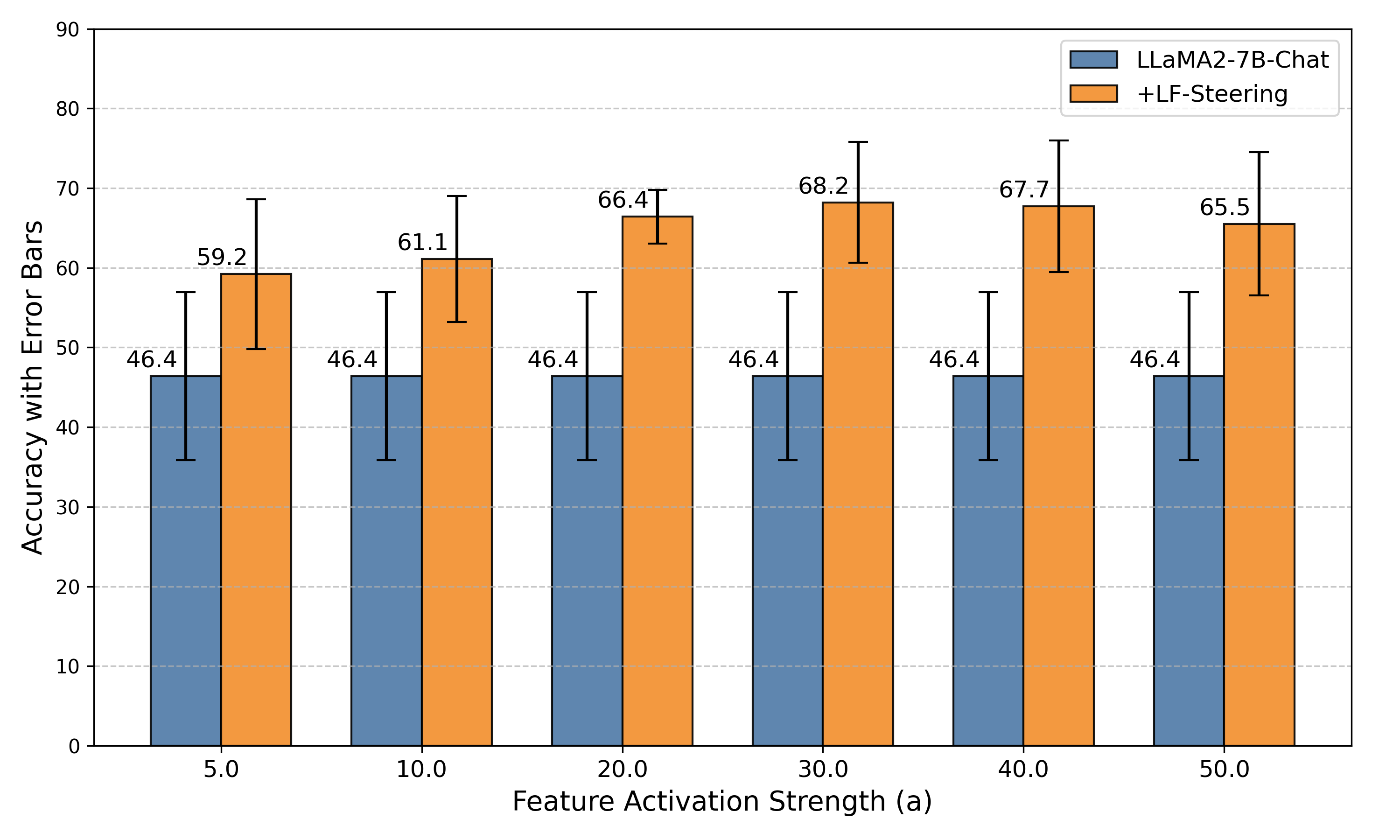}
    \caption{Performance of our proposed activation steering method across different feature activation values.}
    \label{fig:abl_params2}
\end{figure}


We conduct an analysis to assess the impact of the feature activation strength hyperparameter $\alpha$ on the LLM task performance and semantic consistency. We apply our proposed steering method to the LLM and evaluate it using the RobustBOOLQ dataset. The study involves a range of $\alpha$ values selected from the set $\{5.0, 10.0, 20.0, 30.0, 40.0, 50.0\}$. The analysis aims to determine the optimal $\alpha$ value for balancing performance enhancement and maintaining semantic consistency in model steering.

Figure \ref{fig:abl_params2} illustrates that the steered LLM exhibits remarkable robustness across a wide spectrum of $\alpha$ values, ranging from 5.0 to 50.0. Throughout this range, our proposed method consistently enables the backbone LLama2-7B-Chat to improve in both task performance and semantic consistency. However, as anticipated, extreme high value of $\alpha$ (e.g., $\alpha=50.0$) result in performance degradation. In this paper, we set $\alpha$ to a value of $20$.

\subsection{Comparison of the Locality of Semantic Consistency Steering Methods}
To assess the locality \citep{gu2024model} of our proposed steering method and other steering methods, we tested whether these methods might have negative effects on out-of-domain (OOD) datasets. In specific, four OOD datasets were utilized: AG News \citep{Zhang2015CharacterlevelCN} and IMDB \citep{imdb_data} for NLU tasks, and CNN/Daily Mail \citep{see-etal-2017-get} along with XSum \citep{Narayan2018DontGM} for NLG tasks. Accuracy was employed as the evaluation metric for the AG News and IMDB NLU datasets, whereas the ROUGE-L metric \citep{lin2004rouge} was utilized to assess the CNN/Daily Mail and XSum NLG datasets. 

As shown in Tables \ref{tab:ood_nlu} and \ref{tab:ood_nlg}, these steering methods, after enhancing semantic consistency in LLMs, do not negatively impact these OOD datasets, and even lead to improvements on certain datasets. The only exception is ActAdd \citep{DBLP:journals/corr/abs-2308-10248}, which caused a decrease in classification accuracy for AG News from 70\% to 67.20\%. This could be due to ActAdd \citep{DBLP:journals/corr/abs-2308-10248} using only a single contrastive example for steering, resulting in a larger bias from the single one example. Furthermore, among these steering methods, our proposed LF-Steering method exhibits better locality properties compared to the others. This further highlights the advantage of our proposed feature-level steering approach, which offers more precise steering with lower interference.

\begin{table}[htbp]
\centering
\begin{tabular}{@{}lcc@{}}
\toprule
\textbf{Model}  & \textbf{AG\ News} & \textbf{IMDB}  \\ \midrule
LLama2-7B-Chat & 70.00 & 88.60 \\
+ActAdd & 67.20 & 89.60 \\
+CAA & 69.00 & 89.00 \\
+SCS & 70.20 & 89.40 \\
+LF-Steering & \textbf{70.60} & \textbf{89.60} \\ \bottomrule
\end{tabular}
\caption{Comparison of the locality over different steering methods on AG News and IMDB with a subset of 500 instances each.}
\label{tab:ood_nlu}
\end{table}

\begin{table}[htbp]
\centering
\begin{tabular}{@{}lcc@{}}
\toprule
\textbf{Model}  & \textbf{CNN/Daily Mail} & \textbf{XSum}  \\ \midrule
LLama2-7B-Chat & \textbf{21.36} & 14.28 \\
+ActAdd & 21.31 & 14.43 \\
+CAA & 21.35 & 14.44 \\
+SCS & 21.14 & 14.45 \\
+LF-Steering  & 21.00 & \textbf{14.90} \\ \bottomrule
\end{tabular}
\caption{Comparison of the locality over different steering methods using 500 instances from CNN/Daily Mail and XSum datasets.}
\label{tab:ood_nlg}
\end{table}

\section{Conclusion}
In this paper, we present a novel feature-level activation steering approach that maps transformer layer representations to a sparsely activated, higher-dimensional feature space. Our method identifies and adjusts key features that influence semantic consistency, enabling more precise control by decoupling the underlying representations. Experimental results demonstrate that our approach achieves SOTA performances in semantic consistency, resulting in significant task performance gains across a range of NLU and NLG datasets. 

Although our approach achieves significant improvements, we acknowledge that latent representations contributing to semantic inconsistency may span multiple transformer layers in LLMs. Addressing this issue at a single layer may offer only a partial solution, underscoring the importance of a multi-layer activation steering approach. In the furture, we will focus on identifying and steering circuits across multiple transformer layers to enhance the LLM's semantic consistency \citep{elhage2021mathematical,marks2024sparse}.

\bibliographystyle{named}
\bibliography{ijcai25}

\end{document}